\documentclass[conference]{IEEEtran}
\IEEEoverridecommandlockouts
\usepackage{cite}
\usepackage{amsmath,amssymb,amsfonts}
\usepackage{algorithmic}
\usepackage{graphicx}
\usepackage{textcomp}
\usepackage{xcolor}
\usepackage{hyperref}
\usepackage{xurl}
\usepackage{acronym} 

\acrodef{SNN}[SNN]{Spiking Neural Network}
\acrodef{STDP}[STDP]{Spike-timing-dependent plasticity}
\acrodef{DG}[DG]{Dentate Gyrus}
\acrodef{PC}[PC]{Pyramidal Cell}
\acrodef{CA}[CA]{Cornu Ammonis}
\acrodef{LIF}[LIF]{Leaky Integrate-and-Fire}
\acrodef{EC}[EC]{Entorhinal Cortex}
\acrodef{ANN}[ANN]{Artificial Neural Network}

\def\BibTeX{{\rm B\kern-.05em{\sc i\kern-.025em b}\kern-.08em
    T\kern-.1667em\lower.7ex\hbox{E}\kern-.125emX}}

\begin{document}


\title{Spike-based computational models of bio-inspired memories in the hippocampal CA3 region on SpiNNaker}



\author{\IEEEauthorblockN{Daniel Casanueva-Morato, Alvaro Ayuso-Martinez, Juan P. Dominguez-Morales, \\Angel Jimenez-Fernandez and Gabriel Jimenez-Moreno}
\IEEEauthorblockA{\textit{Robotics and Technology of Computers Lab.} \\
\textit{Universidad de Sevilla}\\
Sevilla, España \\
dcasanueva@us.es}
}


\maketitle

\begin{abstract}

The human brain is the most powerful and efficient machine in existence today, surpassing in many ways the capabilities of modern computers. Currently, lines of research in neuromorphic engineering are trying to develop hardware that mimics the functioning of the brain to acquire these superior capabilities. One of the areas still under development is the design of bio-inspired memories, where the hippocampus plays an important role. This region of the brain acts as a short-term memory with the ability to store associations of information from different sensory streams in the brain and recall them later. This is possible thanks to the recurrent collateral network architecture that constitutes CA3, the main sub-region of the hippocampus. In this work, we developed two spike-based computational models of fully functional hippocampal bio-inspired memories for the storage and recall of complex patterns implemented with spiking neural networks on the SpiNNaker hardware platform. These models present different levels of biological abstraction, with the first model having a constant oscillatory activity closer to the biological model, and the second one having an energy-efficient regulated activity, which, although it is still bio-inspired, opts for a more functional approach. Different experiments were performed for each of the models, in order to test their learning/recalling capabilities. A comprehensive comparison between the functionality and the biological plausibility of the presented models was carried out, showing their strengths and weaknesses. The two models, which are publicly available for researchers, could pave the way for future spike-based implementations and applications.
\end{abstract}

\section{Introduction}\label{intro}


Neuromorphic engineering is a concept that was presented by Carver Mead in \cite{mead1990neuromorphic}, and it is a field that focuses on studying, designing and implementing hardware and software with the aim of mimicking the basic principles of biological nervous systems. Its main inspiration comes from studying and replicating how the brain efficiently solves complex problems \cite{indiveri2011neuromorphic}. Based on this biological approach, the information in neuromorphic systems is represented with action potentials (i.e., asynchronous electric pulses, also called spikes) generated by artificial neurons. Spikes are transmitted between different processing layers, leading up to a higher-level functionality. This approach has a clear advantage both in terms of power consumption and real-time capabilities when compared to traditional methods \cite{NeuromorphNature} \cite{zhu2020comprehensiveReview}.


In order to process the aforementioned spiking information, a bio-inspired computational approach is also needed. A specific type of biologically-plausible neural networks called \acp{SNN} are commonly used for this purpose. These are able to process asynchronous spikes instead of floating point or integer numbers, as traditional artificial neural networks would do. \acp{SNN} consists of bio-inspired neuron models, which are interconnected by means of synapses, and mimic the way in which the senses and the brain process the information in living organisms \cite{ahmed2020brain}. In the last years, \acp{SNN} have been used for many different tasks, such as speech recognition \cite{wu2018biologically, dominguez2018deep}, image classification \cite{o2013real, rueckauer2017conversion}, sensory fusion \cite{o2013real, schoepe2020live}, motor control \cite{perez2013neuro, glatz2019adaptive} and bio-inspired locomotion \cite{gutierrez2020neuropod, lopez2021neuromorphic, donati2016novel}, among many others, proving that they could be useful in a wide variety of fields and allowing more efficient and lower power-consumption devices to perform in the same manner as their traditional counterparts.



Among all the different regions within the brain, one of the most interesting ones in terms of its biological behavior is the hippocampus. This component functions primarily as a short-term memory, storing rapid (on-the-fly) and unstructured information coming from the different sensory streams of the cerebral cortex. The hippocampal formation is composed of three main parts: \ac{DG}, the Hippocampus proper (\ac{CA}) and the Subiculum. It is in CA3, a subregion of \ac{CA}, where the recurrent collateral network structure capable of storing information is located \cite{rolls2021brain}. Making use of the mechanisms underlying this sector of the brain to create an efficient and functional bio-inspired memory with \ac{SNN} technology could pave the way for future low-power and low-latency applications.



Previous works that can be found in the literature already proposed different bio-inspired hippocampus memory models. In \cite{tan2013hippocampus}, a hybrid memory architecture is proposed, which combines the structure of CA3 as an oscillating network to repeat sequences of input patterns and the memory structure of the neocortex to store them by modifying its synaptic weights. However, it is only capable of storing simple pattern sequences, and features a neuron model whose oscillation rate is controlled by an external signal from outside the network. In \cite{zhang2016hmsnn}, a model of the hippocampus consisting of \ac{DG} and CA3 is proposed. However, it is not purely spike-based, since the equations that govern CA3 are modeled in digital logic and \ac{DG} is developed as a traditional \ac{ANN}. Other works regarding this topic can be found in the literature, but some are approached from a purely theoretical point of view, such as \cite{shiva2016continuous}; others only perform proofs of concept, such as \cite{oess2017computational}, where a spatial navigation mechanism is evaluated using a static CA3 model with previously stored information in which no learning is applied.



The lack of fully functional bio-inspired spike-based hippocampal memory models that could be used for real-case scenarios including robotics or many other applications was the main motivation for this work. Therefore, two different bio-inspired memory models are proposed: the first model has a closer biological approach, whereas the second model follows a more functional approach. The former is characterised by a constant oscillatory activity in the CA3 attractor network, while the latter adds additional mechanisms on top of the first model to precisely control the network activity. These models have been implemented on the SpiNNaker hardware platform and tested in a series of experiments aimed at demonstrating their performance. The results of the experiments will be used as the basis for analyzing the advantages and disadvantages of moving towards or away from biology in search of functionality.





The main contributions of this work include the following:

\begin{itemize}
    \item Two fully-functional spike-based bio-inspired hippocampal memory models are proposed. These are capable of storing and retrieving complex patterns with certain limitations.
    \item The proposed models were not simply simulated in software, but implemented and emulated on the SpiNNaker hardware platform.
    \item The source code of the proposed models is publicly available, together with all the necessary details regarding the \ac{SNN} architectures.
\end{itemize}
\section{Biological model of the hippocampus and its function as short-term memory}

\begin{figure}[htbp]
    \centerline{\includegraphics[scale=0.30]{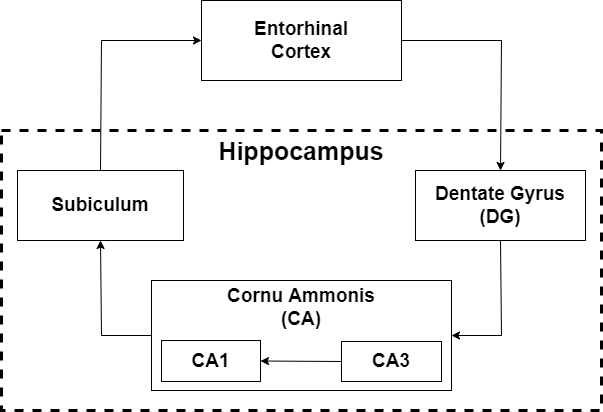}}
    \caption{Diagram of the different layers of the hippocampus and the main connecting pathways between them.}
    \label{hippStruct}
\end{figure}

The hippocampus belongs to the limbic system, the region of the brain that controls emotional responses and is involved in smell, appetite, eating habits, sleep and certain areas of the memory \cite{rajmohan2007limbic}. Specifically, the hippocampus or hippocampal system is a structure inside the brain that consists of the 3 following layers \cite{anand2012hippocampus} \cite{wible2013hippocampal}, as can be seen in Fig.~\ref{hippStruct}:

\begin{itemize}
    
    \item \acf{DG}: it is the input area for the information coming from the \ac{EC}. It plays an important role in the mechanism of pattern separation of the input information; specifically, it is responsible for dispersing the content of the information to achieve a greater degree of orthogonalization, i.e., converting the input information into a representation as different as possible from any other input information. The purpose of this mechanism is to improve both storage capacity and memory access.
    
    
    \item Hippocampus (\ac{CA}): it consists of \acp{PC}, and is divided into CA1, CA2, CA3 and CA4, with CA1 and CA3 being the most relevant in terms of functionality. This region is where the input information is stored and the stored information is recalled and recoded, when requested from the outside.
    
    
    \item Subiculum: it receives the output from the hippocampus and redirects it back to the \ac{EC}.
\end{itemize}

The \ac{EC} (Brodmann's area 28) acts as the main information input and output pathway of the hippocampal system. It has as many input (forward) as output (backward) projections. Through this cortex, the hippocampus receives information from all the different sensory information streams in the brain (including spatial) coming from the neocortex and sends the processed information back to these sensory stream units.


CA3 acts as an autoassociative or attractor memory due to its recurrent collateral network structure, as can be seen in Fig.~\ref{hippStruct}. This region also contains recurrent inhibitory interneurons that regulate the global activity during periods of network oscillation. By receiving all the information from the brain's sensory streams and having the ability to store them in an associated and unstructured way, the hippocampus acts as a short-term memory involved in episodic memory (memory of specific autobiographical events of the past). Therefore, it must be able to store this information (CA3) in a distinct way (separating memories corresponding to different moments thanks to \ac{DG}) and retrieve this information later in order to use it (via CA1 and the subiculum) \cite{rolls2021brain}.

\begin{figure}[htbp]
    \centering
    \centerline{\includegraphics[scale=0.30]{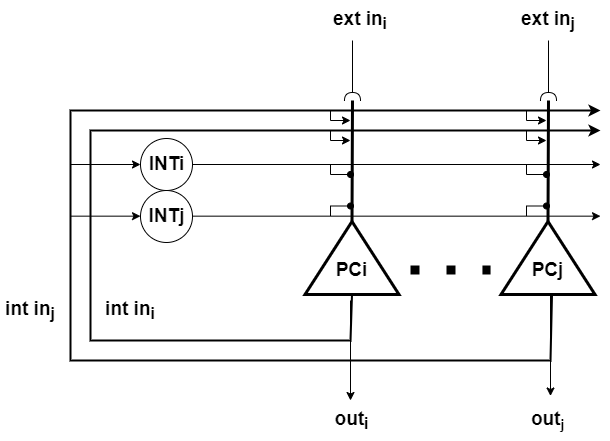}}
    \caption{Internal structure of CA3. It consists of a recurrent collateral network of \acp{PC} with external inputs from the \ac{DG} (ext) and internal inputs from the collateral connections (int), as well as collateral inhibitory inputs from a population of interneurons (INT) that regulate the rate of activity of the network. All the connections are excitatory (lines ending with an arrowhead) except for the projections between the interneurons back to the input of the \acp{PC}, which are inhibitory (lines ending with a circle).}
    \label{ca3struct}
\end{figure}
\section{Materials and Methods}

\subsection{Spiking Neural Networks}


In order to implement bio-inspired memory models based on the hippocampus, third-generation of neural networks, i.e., \acp{SNN}, were used. This kind of networks consists of interconnected neuron models that are not only inspired by biology, but also attempt to mimic their biological counterparts, in order to incorporate the neurocomputational capabilities found in nature \cite{ahmed2020brain}.


As was introduced in section \ref{intro}, \acp{SNN} present an asynchronous event-based functioning and communication thanks to the use of spikes. With the spiking activity, they incorporate the concept of space and time into the neural networks through connectivity and plasticity. It is possible to interpret \ac{SNN} models, i.e., to extract the set of rules that govern them and to understand the learning mechanism \cite{lobo2020spiking}. These event-based networks only have to compute the information when an event occurs (a spike is generated and/or received). This aspect makes \acp{SNN} very efficient from a computational point of view. Moreover, \acp{SNN} present a distributed learning process thanks to the use of \ac{STDP} \cite{caporale2008spike}, which obtains the necessary information from the electrical impulses encoded between local neurons to define the weight of the synapses, i.e., to regulate the learning of the network \cite{ahmed2020brain}.


\acp{SNN} present key features for being implemented in hardware compared to traditional \acp{ANN}, allowing for higher efficiency and lower power consumption: multiplications are replaced with adders and shifts, the information that is transmitted between neurons is only 1-bit in size instead of integer or floating point numbers, etc. \cite{lobo2020spiking}.

Different hardware platforms particularly designed for implementing and simulating \acp{SNN} can be found in the literature. Some of the most well-known ones are SpiNNaker \cite{furber2014spinnaker}, Loihi \cite{davies2018loihi} and TrueNorth \cite{merolla2014million}. In this work, we used SpiNNaker as the hardware platform in which the different \ac{SNN} models presented were implemented and emulated.

\subsection{SpiNNaker}

SpiNNaker \cite{furber2014spinnaker} is a massively-parallel multi-core computing system, which was designed for being able to model very large \acp{SNN} in real time. Each SpiNNaker chip consists of 18 general-purpose ARM968 cores, running at 200 MHz, which communicate the information by means of packets carried by a custom interconnect fabric \cite{furber2013overview}. In this work, a SpiNN-5 machine was used. SpiNN-5 is a 48-node circuit board and, thus, has 864 ARM processor cores, which are commonly deployed as 768 application cores, 48 Monitor Processors and 48 spare cores. A 100 Mbps Ethernet connection is used as an I/O interface and for sending scripts and commands to the board. A custom software package called sPyNNaker \cite{rhodes2018spynnaker} allows running PyNN \cite{davison2009pynn} simulations directly on the SpiNNaker board. This makes the platform very straight-forward to work with, since all the codes regarding the design and implementation of \acp{SNN} can be done using high-level functions described in Python programming language.

\section{Hippocampus-based computational models of memory}
\label{modelos}


This paper presents two bio-inspired computational memory models of the hippocampus with different degrees of closeness to the theoretical bases. Before describing all the implementation details of these models, it is necessary to define the neuron models, synapses and learning rules used first.


Starting with neurons, many different models can be found in the literature, each of them with specific characteristics depending on the level of abstraction. Among them, the most widely-used in the field of \acp{SNN}, and the one that has been used for the proposed model implementations is the standard \ac{LIF}. This model describes the behavior of a neuron as an RC electrical circuit whose potential value is either increased or decreased with the arrival of input spikes and, after reaching a threshold value, generates output spikes. For both of the implemented spike-based memory models, the parameters of the \ac{LIF} neurons used are: $c_{m}$: 0.27 nF, $\tau_{m}$: 10.0 ms, $\tau_{synExc}$: 0.3 ms, $\tau_{synInh}$: 0.3 ms, $v_{reset}$: -60.0 mV, $v_{rest}$: -60.0 mV, $v_{thresh}$: -55.0 mV, where $c_{m}$ is the capacitance, $\tau_{m}$ is the time-constant of the RC circuit, $\tau_{synExc}$ and $\tau_{synInh}$ are the excitatory and inhibitory input current decay time-constants, respectively, $v_{reset}$, $v_{rest}$ and $v_{thresh}$ are the membrane potentials at which the neuron is set immediately after generating a spike, during the inactivity period and the threshold at which the neuron spikes, respectively. The value of the $\tau_{refrac}$ parameter, which refers to the refractory period of the neuron (the time that the neuron is disabled after generating a spike and stops integrating any input spike) is set different for the two \ac{SNN} models proposed.


Synapses can be modelled as a weighted edge in a graph connecting a source (pre-synaptic) neuron to a target (post-synaptic) neuron. The learning mechanism of \acp{SNN} focuses on the plasticity of the synapses, i.e., the ability to create, remove or modify the weight of the synapses. This weight represents an increase or a decrease in the action potential that will be applied to the postsynaptic neuron when a spike is received. Therefore, a modification of the weight would pose an influential change.


Among all the different learning mechanisms where the concept of synapse plasticity is considered, the most widely-known and the one that was used in this work is \ac{STDP}. It is based on a Hebbian learning mechanism in which the weights of synapses are modified in proportion to the temporal correlation between pre- and post-synaptic neuronal activity \cite{sjostrom2010spike}. The weight of a synapse will increase whenever a presynaptic spike is received before the postsynaptic spike, and will decrease otherwise, by an amount proportional to the time difference between the two spikes (the closer in time, the greater). \ac{STDP} is implemented in SpiNNaker, although it presents some particularities: the evaluation of the weight change (sum of cumulative increments and decrements) in a synapse is produced only on presynaptic spikes \cite{jin2010implementing}. 

Based on the materials and methods described, two different CA3 models were proposed and implemented: an oscillatory CA3 model (Section \ref{subsec:oscillatoryCA3}), and a regulated CA3 model (Section \ref{subsec:regulatedCA3}).

\subsection{CA3 memory model with oscillatory activity}
\label{subsec:oscillatoryCA3}

\begin{figure}[htbp]
    \centerline{\includegraphics[scale=0.30]{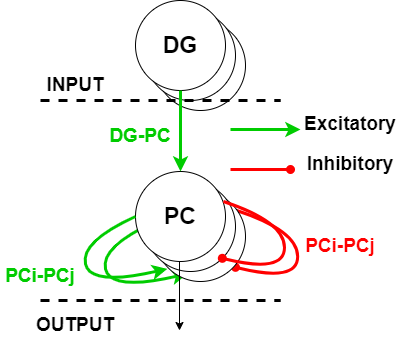}}
    \caption{Diagram of the DG-CA3 network with oscillatory activity in the recurrent collateral connections of CA3. Excitatory synapses are shown in green and inhibitory synapses in red. The excitatory synapses where \ac{STDP} is applied are $PC_i-PC_j$.}
    \label{ca3oscstruct}
\end{figure}

The first of the two models implemented (see Fig.~\ref{ca3oscstruct}) is the closest to the biological behavior, as described in the literature, i.e., an attractor network that maintains an internal state and is constantly oscillating. The design consists of a layer of \ac{DG} neurons that act as input to the system and a layer of \acp{PC} that form CA3. \ac{DG} has one-to-one excitatory connections with \ac{PC} neurons, and \ac{PC} neurons are interconnected between them by means of one-to-all recurrent collateral connections, both excitatory and inhibitory. The excitatory collateral synapses between \ac{PC} neurons are where \ac{STDP} is applied, and they are responsible for the storage of patterns in the network. The rest of the synapses are static.

The static synapses of the network are set with a delay of 1 ms. Regarding the weights, the recurrent inhibitory connections ($PC_i-PC_j$) are set with a weight of 1.5 nA, and the connections between \ac{DG} and \ac{PC} neurons are set to a value that is high enough for a presynaptic spike to always generate a postsynaptic spike. As for the \ac{STDP} learning rule, the parameters used were: $\tau_plus$ = 3.0 ms, $\tau_minus$ = 2.0 ms, $A_plus$ = 6.0, $A_minus$ = 3.0, maximum weight = 12.0 nA, minimum weight = 0.0 nA, initial weight = 0.0 nA and delay = 1.0 ms, where $\tau_plus$ and $\tau_minus$ are the decay time-constants that control the amount of weight increase or decrease, respectively, and $A_plus$ and $A_minus$ define the maximum weight to respectively add during potentiation or subtract during depression.


This network has two main phases: a learning phase and a recall phase. In the former, the patterns to be learned are introduced to the network. The \ac{STDP} rule changes the weights of the synapses and, thus, the internal state of the attractor. In the second phase, a part of the pattern to be recalled (cue) is introduced as input. After a few oscillation cycles, the network returns the rest of the pattern. Once the first cue arrives at the network, it starts oscillating and it is the passing of the cue that makes it oscillate from one state to another, with the help of the lateral inhibitions, where the state is the pattern to be remembered. For this to happen, the neurons need to be configured with a refractory period of 0, so that they will be constantly oscillating between states.

\subsection{CA3 memory model with regulated periods of activity}
\label{subsec:regulatedCA3}

\begin{figure}[htbp]
    \centerline{\includegraphics[scale=0.25]{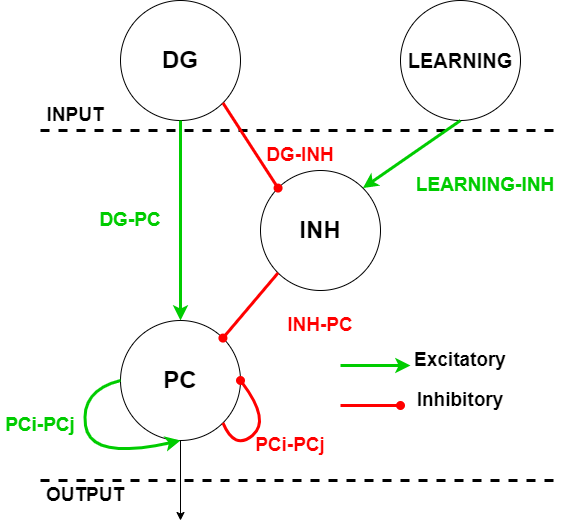}}
    \caption{Diagram of the DG-CA3 network with regulated activity to avoid constant oscillation. Excitatory synapses are shown in green and inhibitory synapses in red. The excitatory synapses where \ac{STDP} is applied are $PC_i-PC_j$. It has an additional mechanism with inhibitory interneurons (INH) to regulate the activity in the learning phase.}
    \label{ca3pcstruct}
\end{figure}

The second model implemented (see Fig.~\ref{ca3pcstruct}) is an extension of the previous model, where an activity inhibition mechanism is added, changing the refractory period from 0 to 2 ms and using stronger lateral inhibitions that will depend on the number of \ac{PC} neurons. These changes aim to precisely regulate and control the internal activity of the CA3 recurrent collateral network, reducing power consumption and achieving greater stability.

The inhibition mechanism consists of a layer of inhibitory interneurons (INH) of the same size as the \ac{DG} and the \ac{PC} layer, and receives two inputs: an excitatory signal indicating whether the network is in the learning phase (LEARNING), and an inhibitory input with one-to-one connectivity from the \ac{DG} (DG-INH). The output of the network is sent to \ac{PC} neurons via inhibitory one-to-one synapses (INH-PC). The neuron models and static synapses are identical to those used for the DG-PC connections. This inhibitory mechanism mimics the inhibitory interneurons that are present in some biological models of the hippocampus in order to regulate the oscillation rate of the network only while learning, but not during the recall phase. This approach is also bio-inspired, but it is not as close to biology as the previous model, mainly due to the introduction of the LEARNING signal.


The modifications performed in this model with respect to the first model allow having a precise event-based activity network, rather than a constant oscillation, with the consequent advantages and disadvantages that are detailed and discussed in section \ref{resultados}. It also has a learning phase and a recall phase in which the \ac{STDP} remains static. This means that the weights obtained after the learning phase are used in the same network but with static synapses to perform the recall without the problem of forgetting data due to it. 
\section{Experimentation and results}\label{resultados}



In order to test the behavior of the models implemented in SpiNNaker, different experiments were performed. The first set of experiments were focused on verifying the performance of the proposed models using completely orthogonal patterns (patterns that are completely different from each other and have nothing in common) as input, which is the minimum requirement for the memory to be functional. Then, the second set of experiments were focused on testing the models when using non-orthogonal patterns (patterns that have some parts in common) as input. For all the experiments that were performed, networks with 15 \ac{PC} neurons were used.

\subsection{CA3 memory model with oscillatory activity}

\begin{figure*}[htbp]
    \centerline{\includegraphics[width=.90\linewidth]{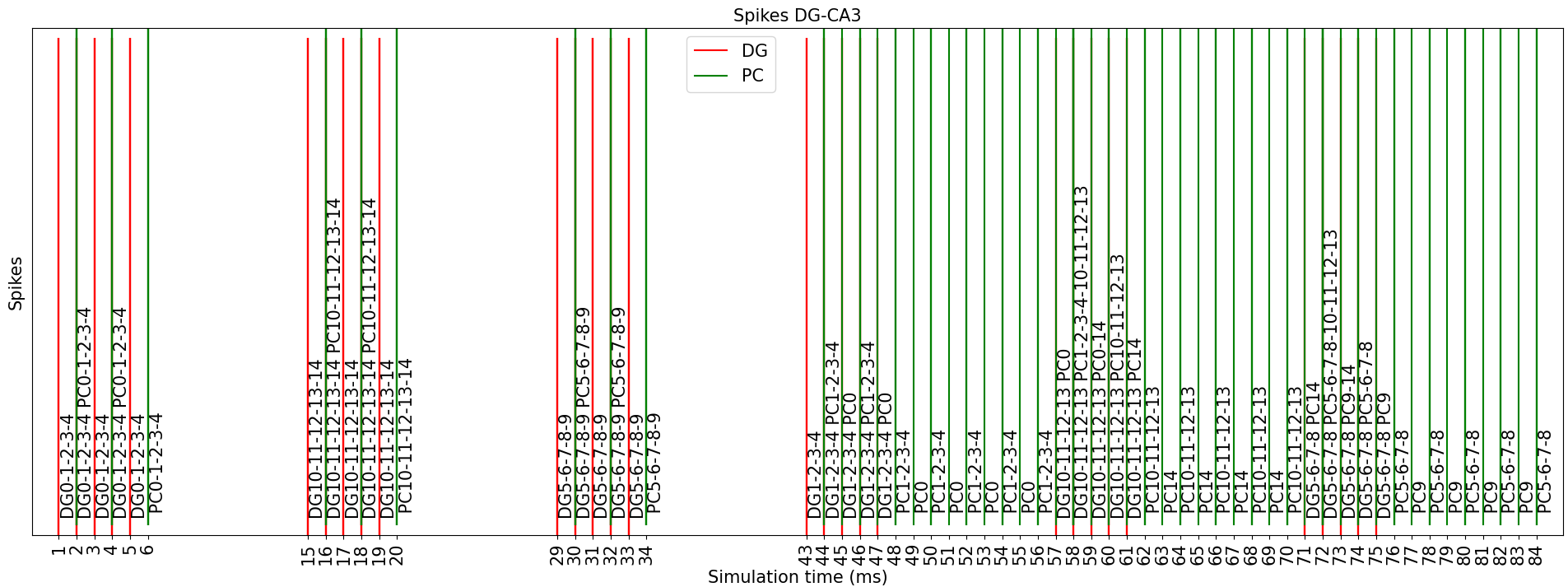}}
    \vspace{.5cm}
    \centerline{\includegraphics[width=.90\linewidth]{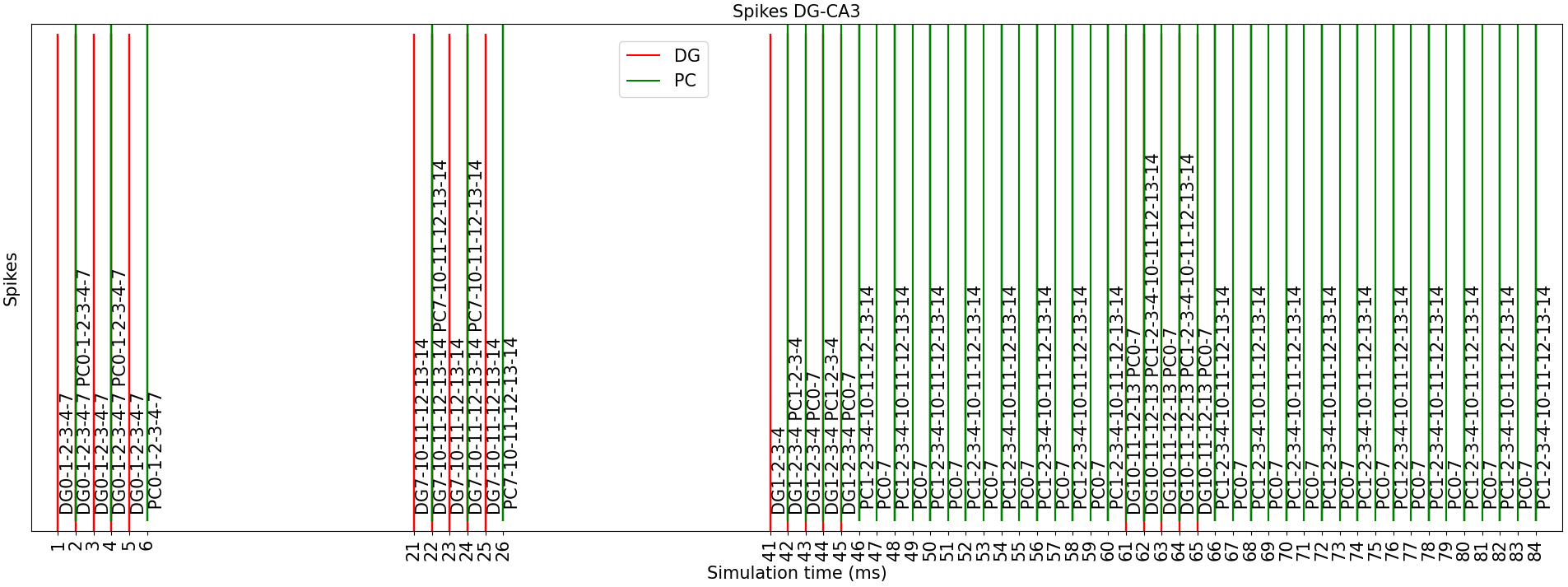}}
    \caption{Results of the experiments performed using the oscillatory network for the learning and recall of 3 orthogonal patterns (top) and 2 non-orthogonal patterns (bottom). During the first half of both simulations, the learning phase of the patterns can be observed, while the second half corresponds to the recall phase of the different patterns.}
    \label{experimentca3osc}
\end{figure*}

In the first set of experiments, the oscillatory network was able to learn and correctly recall different combinations of 2-4 fully orthogonal patterns. In the top plot of Fig.~\ref{experimentca3osc}, the results for the case with 3 orthogonal patterns can be observed. In the first 6 milliseconds (milliseconds 1-6) the pattern 0-1-2-3-4 was used as input to the network, simulating the arrival of those 5 spikes coming from the first 5 \ac{DG} neurons. The input was introduced to the network up to 5 consecutive times (milliseconds 1-5), where, thanks to \ac{STDP}, the synaptic weights are modified to learn the spike pattern, also producing an output in the \ac{PC} neurons corresponding to the stored pattern. The same process is done for patterns 10-11-12-13-14 and 5-6-7-8-9 at milliseconds 15-20 and 29-34, respectively. After learning the patterns during the learning phase, the 1-2-3-4 cue (milliseconds 43-47) is introduced through \ac{DG}, and the recurrent collateral network of \acp{PC} starts to oscillate until the rest of the first pattern (0) is recalled (millisecond 49) reliably (it also appears at milliseconds 45 and 47 while the input sample is being given; therefore, it is not reliable). The oscillatory activity recalling the first pattern is maintained until the arrival of the cue corresponding to the second pattern (10-11-12-13) in milliseconds 57-61, which later allows obtaining the rest of it (14) in millisecond 63. The network continues oscillating after that, but this time recalling the second pattern, and the same process is repeated for the third pattern with cue 5-6-7-8 in milliseconds 71-75, obtaining in millisecond 77 the rest of it (9). As can be seen, once it starts oscillating in a state, it stays in that state until another cue is given. The recalling process has been illustrated with a cue of 4 components of the pattern to recall the fifth component; however, the same behavior would happen when using any combination of any size of the pattern as a cue.


Although the experiments were performed storing up to 4 patterns, the storage capacity of the network is determined by the number of \ac{PC} neurons and the size of the patterns. Thus, a network with 20 \acp{PC} and input patterns of size 4 would be able to store up to 5 orthogonal patterns. For this to work, it is necessary that the learning and recalling phases are separated, since once the first recall cue is used as input, the network starts to oscillate continuously. Therefore, if learning takes place during oscillation, the network will learn the sum of the input patterns together with the current state of the network instead of the original input patterns. Furthermore, for the \ac{STDP}-based learning process to work correctly, it is necessary to present the pattern several times and achieve constant positive reinforcement on the appropriate connections and negative reinforcement on the others. In this model, it is necessary to only present each of the input patterns 5 times in a row for it to be learned correctly.


Given the architecture of the network, the oscillation between states causes the new state to be reinforced and the previous one to be forgotten when changing from one state to another. Therefore, each pattern can only be remembered once, i.e., the model is a volatile memory. This makes sense from a biological point of view: the hippocampus is a short-term memory, and thus, information is learned and forgotten fairly quickly, being retained longer as long as the network keeps recalling it, or forgotten if the network changes to a different state because the information is no longer useful or simply stored in another memory region after being processed.



Regarding experimentation with non-orthogonal patterns (bottom plot of Fig.~\ref{experimentca3osc}), the network is able to learn them correctly. However, when it comes to recalling them, due to the oscillatory nature of the network, it will start remembering the pattern correctly, but then the common part with other patterns will be used as a cue of the other patterns that have it. Then, the network will oscillate to a new state of addition of all these patterns, with the consequent learning of this association of patterns and the forgetting of the patterns individually. From a biological point of view, it makes sense in some way, since anterior regions of the hippocampus have the goal of separating patterns through a dispersion of information (increasing the degree of orthogonality), meaning that CA3 also has problems working with non-orthogonal patterns.

\subsection{CA3 memory model with regulated periods of activity}

\begin{figure*}[htbp]
    \centerline{\includegraphics[width=.90\linewidth]{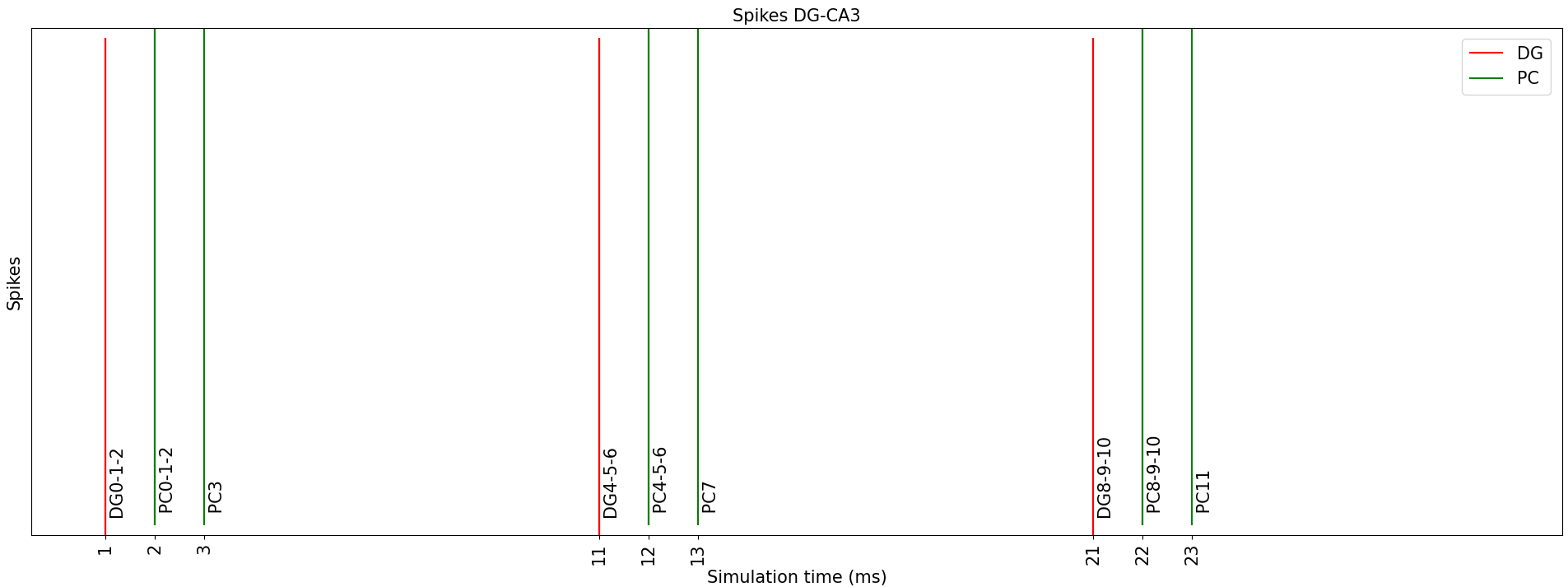}}
    \vspace{.5cm}
    \centerline{\includegraphics[width=.90\linewidth]{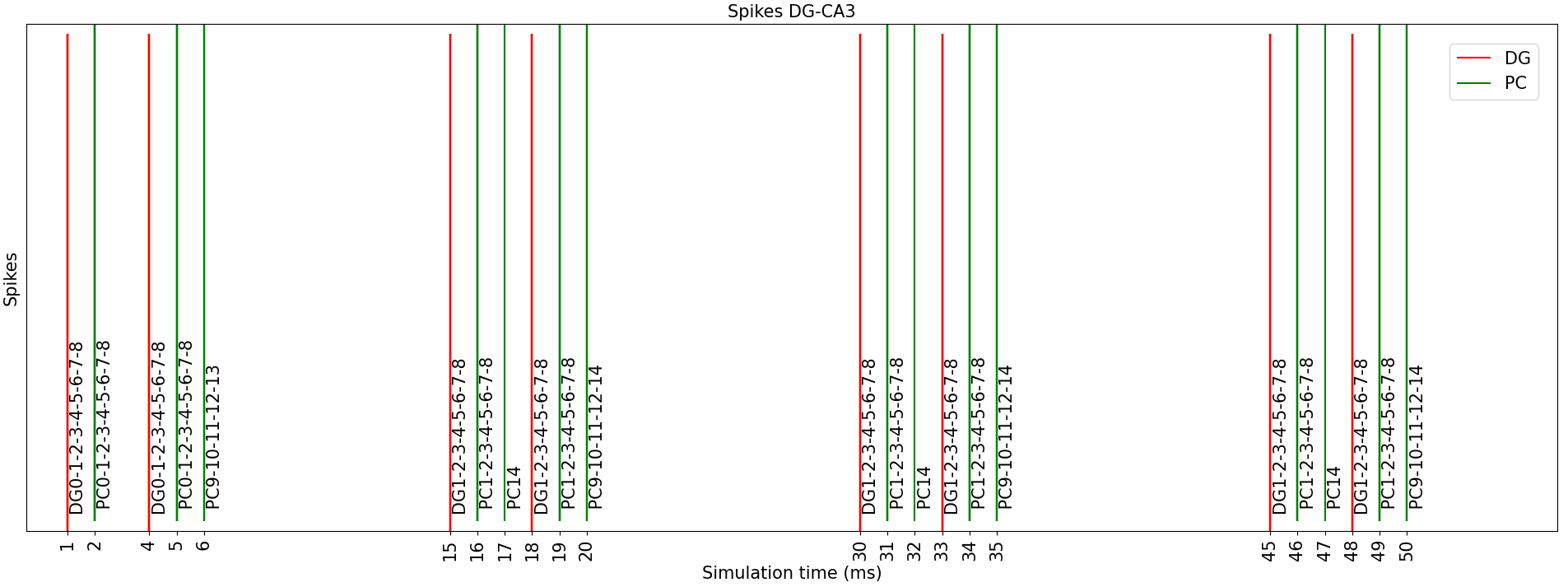}}
    \caption{Results of the experiments performed using the network with regulation of activity periods for the learning and recalling of 3 orthogonal patterns (top) and 2 non-orthogonal patterns (bottom). During the first half of both simulations, the learning phase of the patterns can be observed, while the second half corresponds to the recall phase of the different patterns.}
    \label{experimentca3pc}
\end{figure*}

\begin{table*}[]
    \centering
    \caption{Comparison in terms of resources, performance and functionality of both memory models.}
    \label{tbcomp}
    \resizebox{\textwidth}{!}{%
    \begin{tabular}{l|ccccccccc}
    Network           & \multicolumn{1}{l|}{Neurons} & \multicolumn{1}{l|}{Static synapses} & \multicolumn{1}{l|}{STDP synapses} & \multicolumn{1}{l|}{Learning (ms)} & \multicolumn{1}{l|}{Recall (ms)} & \multicolumn{1}{l|}{Operations phases} & \multicolumn{1}{l|}{Patterns} & \multicolumn{1}{l|}{Energy consumption} & \multicolumn{1}{l|}{Stability and Reliability} \\ \hline
    Oscillatory DG-CA3 & 2*n                          & n                                    & n*(n-1)                            & 14                                 & 14                               & Both in same simulation                & Only orthogonal               & High                                    & Low                                            \\
    Regulated DG-CA3  & 3*n + 1                      & 4*n                                  & n*(n-1)                            & 50                                 & 14                               & One simulation per phase               & Both                          & Low                                   & High                                          
    \end{tabular}%
    }
\end{table*}


A set of experiments for learning and recalling different combinations between 2 and up to 4 fully orthogonal patterns were carried out on the network. In the upper plot of Fig.~\ref{experimentca3pc}, the result of the network regarding the recall phase of 3 orthogonal patterns can be observed. For the sake of clarity, the learning process of the patterns in this model has not been included, since it is also similar to the one that can be seen in the experiments that were performed for the oscillatory model. In the recall phase, the weights were fixed to a weight value obtained from the result of the previous training phase (using \ac{STDP}). In this case, each pattern only needed to be given to the network a total of 4 times in order to learn it appropriately. The learned patterns were 0-1-2-3, 4-5-6-7 and 8-9-10-11, and the cues provided were 0-1-2, 4-5-6 and 8-9-10. The patterns were recalled correctly (the remaining part of the pattern was obtained for each cue: 3, 7 and 11, respectively) by only giving each cue once to the network. The network did not need any kind of oscillation to be able to return the remaining part of each pattern, but directly recalled the appropriate result, thus, achieving optimum performance in terms of power consumption.


As with the previous model, up to 4 patterns were tested, but the storage capacity of the network depends on the number of \ac{PC} neurons and the size of the orthogonal patterns.


Regarding the experiments with non-orthogonal patterns, the network was able to learn and recall them correctly. In the learning phase of the bottom part of Fig.~\ref{experimentca3pc}, which is not represented in the figure, 2 non-orthogonal patterns were shown 4 times: 1-2-3-4-5-6-7-8-9-10-11-12-14 (all but 0 and 13) and 0-1-2-3-4-5-6-7-8-9-10-11-12-13 (all but 14). After learning, the recall of both patterns was performed using the following cues: 1-2-3-4-5-6-7-8 and 0-1-2-3-4-4-5-6-7-8. After providing the cue containing common parts to both patterns two times, the network was able to correctly recall the remaining part of each of them: 9-10-11-12-14 and 9-10-11-12-13.


Despite working correctly with non-orthogonal patterns, the main problem of this network is that it needs a very precise configuration of the recurrent lateral inhibition weights to perform correctly. This configuration is complex to tune automatically, since it depends not only on the size of the network, the number of patterns to store and their percentage of non-orthogonality, but also on other internal factors of the network such as the resulting weights of the \ac{STDP} connections, among  others, which are difficult to model with an equation. Therefore, the proper use of this model requires a prior configuration step to ensure that it works correctly under the desired conditions.


Thanks to the regulated behavior of the network, which allows having only the necessary activity over time, this model is not only more energy efficient (it only processes the information when there is a new input) but also capable of recalling any internally stored pattern as many times as desired without forgetting it over time or after recalls (which happened in the first model), regardless of whether it is orthogonal or not.
\section{Discussion and comparison of the models}


In terms of the architecture and resources, the model with controlled activity is an extension of the first model; therefore, it needs more resources, although both scale linearly as a function of \textit{n} (the size of the memory), thus the differences are not substantial between both models, as can be seen in Table~\ref{tbcomp}.


In terms of functionality and performance, based on the results of the experiments, the oscillatory model is more bio-inspired due to its constant activity within the attractor network once it starts to recall. However, this also makes it more energetically expensive and unstable as it is constantly generating spikes, even when there is no operation in progress. The second model has additional regulation mechanisms that allow it to maintain minimal activity (only when performing an operation), which makes it more efficient, stable and reliable, since the same patterns can be recalled several times without being erased from memory. On the other hand, these mechanisms make this model have a higher latency in the learning process, as it requires a certain temporal separation between pattern samples. A comprehensive comparison between both models is presented in Table~\ref{tbcomp}, where learning and recalling times refer to the time it takes from the start of the operation until the timestep when the next operation can be performed (not the time that these operations need, which is shorter).


Finally, the first model does not require a prior configuration and it is able to recall any learned pattern immediately after learning it. However, it does not perform well with non-orthogonal patterns, unlike the second model, which is able to work with both kinds of patterns but requires a prior complex parameter setting.
\section{Conclusions}


In this paper, two bio-inspired memory models of the hippocampus with different levels of abstraction have been proposed, implemented and tested. The difference between the models and their biological counterparts allowed exploring alternative ways of storing the information and illustrate the relationship between plausibility and functionality. To the best of the authors' knowledge, the CA3 models presented in this work are the first to be designed and implemented using \acp{SNN} on the SpiNNaker hardware platform, not simply simulated in software. Moreover, these models, particularly the second one, are fully functional and completely feasible for their use in real applications based on this spiking paradigm, as they are able to learn and recall patterns of different sizes.


The design of the models and the experiments that were performed illustrate an inverse relationship between plausibility and functionality. Although both are fully functional, the first model is more bio-inspired but is less suitable for a general-purpose domain as it has greater instability, reliability and power consumption, as well as the inability to work with non-orthogonal patterns. Certain changes were needed to make it somewhat less bio-inspired in order to achieve a more functional and useful model for its use in real applications, although this involved having greater learning times.  


The strengths and the weaknesses of both models were demonstrated, opening future lines of research in order to study possible improvements. As an example for the first model, the use of collateral recurrent inhibitory interneurons that reduce or even remove the oscillatory activity could be studied, so that it will only remain active for a specific time period when operations are being performed. For the second model, eliminating the need for prior adjustment of the network by modifying or adding connections or populations of neurons could be beneficial, even if the model moves away from plausibility.

The source code regarding the implementation of both hippocampal bio-inspired memories with \acp{SNN} on SpiNNaker is available on GitHub\footnote{\url{https://github.com/dancasmor/Spike-based-computational-models-of-bio-inspired-memories-in-the-hippocampal-CA3-region-in-SpiNNaker}}.
\section*{Acknowledgments}
This research was partially supported by the Spanish grant MINDROB (PID2019-105556GB-C33/AEI/10.13039/501100011033).

Daniel~Casanueva-Morato was supported by a "Formaci\'{o}n de Profesor Universitario" Scholarship from the Spanish Ministry of Education, Culture and Sport.

\bibliographystyle{IEEEtran}
\bibliography{bibliography}

\end{document}